\documentclass[10pt,twocolumn,letterpaper]{article}

\usepackage{cvpr}
\usepackage{times}
\usepackage{epsfig}
\usepackage{graphicx}
\usepackage{amsmath}
\usepackage{amssymb}

\usepackage{array,multirow,graphicx}

\usepackage[breaklinks=true,bookmarks=false]{hyperref}

\usepackage[belowskip=-10pt,aboveskip=5pt]{caption}
\cvprfinalcopy 

\def\cvprPaperID{****} 

\ifcvprfinal\fi
\begin{document}

\title{Semantically Multi-modal Image Synthesis}

\author{Zhen~Zhu$^{1}$\thanks{~Equal contribution} , Zhiliang Xu$^{1}$\footnotemark[\value{footnote}] , Ansheng You$^{2}$, Xiang~Bai$^1$\thanks{~Corresponding author}\\
	$^1${\em Huazhong University of Science and Technology}, $^2${\em Peking University}\\ 
	{\tt \small \{zzhu, zhiliangxu1, xbai\}@hust.edu.cn}, {\tt \small youansheng@pku.edu.cn}
}

\maketitle
\thispagestyle{empty}


\begin{abstract}
	In this paper, we focus on \textbf{semantically multi-modal image synthesis} (SMIS) task, namely, generating multi-modal images at the semantic level. Previous work seeks to use multiple class-specific generators, constraining its usage in datasets with a small number of classes. We instead propose a novel Group Decreasing Network (GroupDNet) that leverages group convolutions in the generator and progressively decreases the group numbers of the convolutions in the decoder. Consequently, GroupDNet is armed with much more controllability on translating semantic labels to natural images and has plausible high-quality yields for datasets with many classes. Experiments on several challenging datasets demonstrate the superiority of GroupDNet on performing the SMIS task. We also show that GroupDNet is capable of performing a wide range of interesting synthesis applications. Codes and models are available at: \url{https://github.com/Seanseattle/SMIS}.
\end{abstract}

\section{Introduction}
\begin{figure}[ht]
	\centering
	\includegraphics[width=0.43\textwidth]{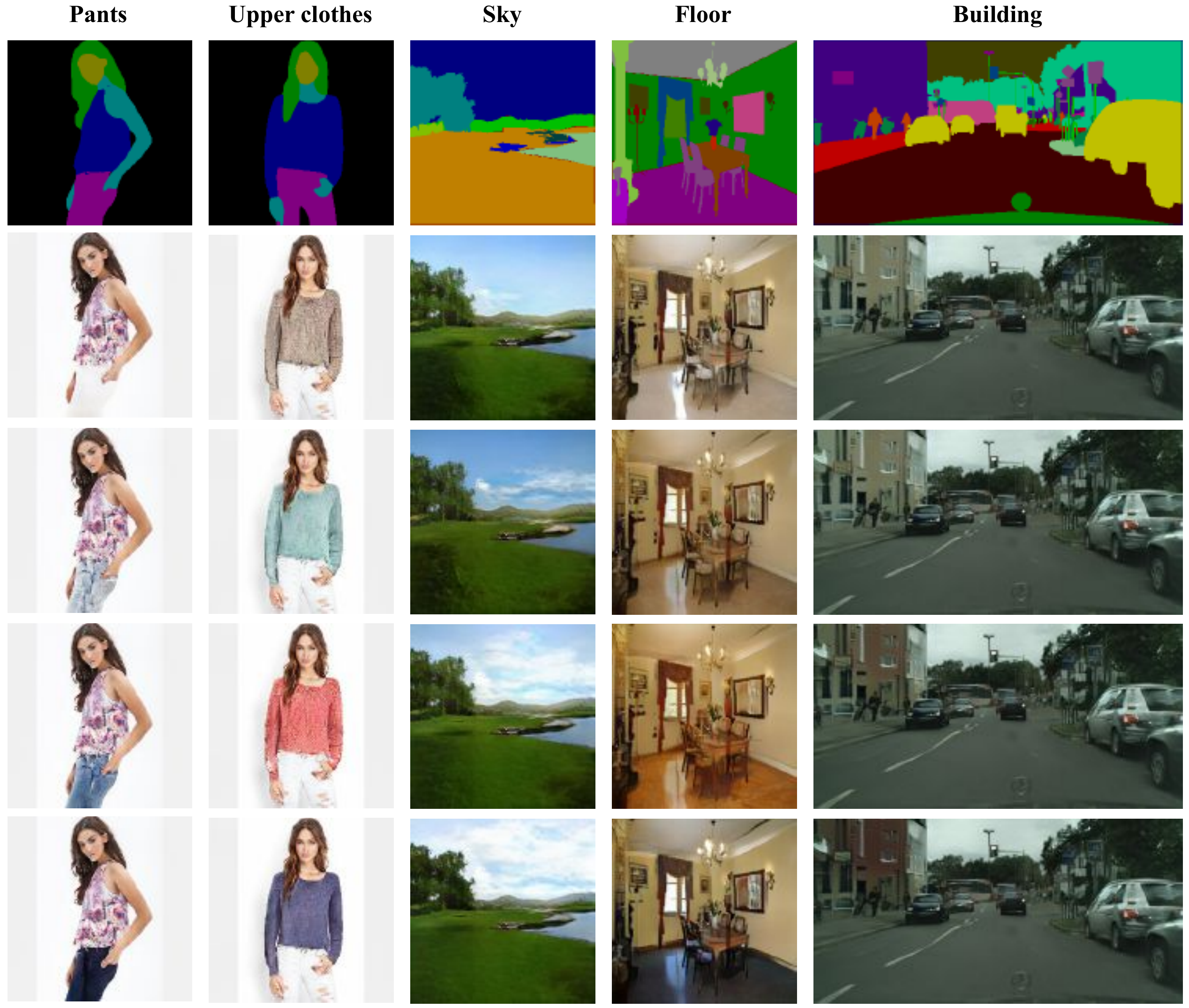}
	\caption{Demonstration of the semantically multi-modal image synthesis (SMIS) task. Text of each column above the images indicates the semantic area that is changing across the whole column. The first row represents the input labels and the rest rows are generated images by our method.}
	\label{fig:teaser}
\end{figure}


Semantic image synthesis, namely translating semantic labels to natural images, has many real-world applications and draws much attention from the community. It is innately a one-to-many mapping problem. Countless possible natural images correspond to one single semantic label. Previous works utilized different strategies for the task: adopting the idea of variational auto-encoder~\cite{SPADE,BiCycleGAN,cVAE-GAN,compatible}, introducing noise while training~\cite{pix2pix}, building multiple sub-networks~\cite{MGPE} and including instance-level feature embeddings~\cite{pix2pixHD}, \etc. While these methods made exceptional achievements in improving image quality and extending more applications, we take a step further to particularly focus on a specific multi-modal image synthesis task that adds more flexibility to control the generated results.


Just imagine a content creation scenario from a human parsing map. With the help of semantics-to-image translation models, the parsing map can be converted to a real person image. It looks good in general, but the upper clothes do not suit your taste. Then comes the problem---either these models do not support multi-modal synthesis, or when these models change the upper clothes, other parts vary accordingly. Neither of these fulfills your intention. To conclude, this user controllable content creation scenario can be interpreted as performing a task that produces multi-modal results at the semantic level with other semantic parts untouched. 
We summarize this task as: \textbf{Semantically Multi-modal Image Synthesis} ({SMIS}). As exemplified in Fig.~\ref{fig:teaser}, for each semantics, we have its specific controller. By adjusting the controller of a specific class, only the corresponding areas are changed accordingly. 

An intuitive solution for the task is to build different generative networks for different semantics and then produce the final image by fusing the outputs of different networks. It is quite similar to the overall scheme of \cite{MGPE}, which focused on portrait editing. However, this type of methods soon face degradation in performance, a linear increase of training time and computational resource consumption under a growing number of classes. 

To make the network more elegant, we unify the generation process in only one model by creatively replacing all regular convolutions in the generator with group convolutions. Our strategy is mathematically and functionally equivalent to \cite{MGPE} when the group numbers of the convolutions are equal to that of the classes. Another strategy we adopt, however, set our paths differently---we decrease the number of groups in the decoder convolutions in the forwarding progress. We observe that different classes have internal correlations among each other, \eg, the color of the grass and the tree leaves should be highly similar. In this case, progressively merging the groups gives the model enough capacity to build inter-correlations among different classes, and consequently improves the overall image quality. Besides, this strategy also considerably mitigates the computation consumption problem when the class number of the dataset is substantial~(\textit{e.g.}, ADE20K~\cite{ADE20K}). We call the generator equipped with these two strategies as \textbf{Group Decreasing Network} (GroupDNet). To evaluate GroupDNet's performance on the SMIS task, we propose two new metrics called mean Class-Specific Diversity (mCSD) and mean Other-Class Diversity (mOCD). The mCSD metric tends to hold high value and the mOCD tends to be low when some semantic parts vary drastically but other parts stay unchanged.


We conduct experiments on several challenging datasets: DeepFashion \cite{DeepFashion},  Cityscapes \cite{Cityscapes}, and ADE20K \cite{ADE20K}. The results show that our GroupDNet introduces more controllability over the generation process, and thus produces semantically multi-modal images. Moreover, GroupDNet maintains to be competitive with previous state-of-the-art methods in terms of image quality, exhibiting the superiority of GroupDNet. Furthermore, GroupDNet introduces much controllability over the generation process and has a variety of interesting applications such as appearance mixture, semantic manipulation, and style morphing.


\section{Related work}


\vspace{0.5ex}\noindent \textbf{Generative models.}~Generative Adversarial Networks (GANs)~\cite{GAN}, comprised of a generator and a discriminator, have the amazing ability to generate sharp images even for very challenging datasets.~\cite{GAN,ProgressiveGAN,BigGAN,StyleGAN}. Variational auto-encoder~\cite{VAE} contains an encoder and a decoder, and requires the latent code yield by the encoder to conform to Gaussian distribution. Its results usually exhibit large diversity. Some methods~\cite{cVAE-GAN} combine VAE and GAN in their models, producing realistic while diverse images. 

\vspace{0.5ex}\noindent \textbf{Conditional image synthesis.}~Conditional Generative Adversarial Networks~\cite{CGAN} inspire a wide range of conditional image synthesis applications, such as image-to-image translation~\cite{pix2pix,MUNIT,pix2pixHD,CycleGAN,DIRT,UNIT,hsim}, super resolution~\cite{SRGAN,perceptualloss}, domain adaption~\cite{CyCADA,curriculum}, single model image synthesis~\cite{NonStaionary,SinGAN,PGIM,InGAN}, style transfer~\cite{AdaIN,Gatys2016,perceptualloss}, person image generation~\cite{poseguided,PATN,compatible} and image synthesis from text~\cite{StackGAN,StackGAN++}~\etc. We focus on transforming conditional semantic labels to natural images while adding more diversity and controllability to this task at the semantic level.

\vspace{0.5ex}\noindent \textbf{Multi-modal label-to-image synthesis.}~There has been a number of works~\cite{shapeandcontext,IMLE,exampleguided} in multi-modal label-to-image synthesis task. Chen~\emph{et. al.}~\cite{PhotographicImageSynthesis} avoided using GAN and leveraged cascaded refinement network to generate high-resolution images. Wang~\emph{et. al.}~\cite{pix2pixHD} added additional instance-level feature channels to the output of the encoder that allows object-level control on the generated results. Wang~\emph{et. al.}~\cite{exampleguided} used another source of images as stylish examples to guide the generation process. Park~\emph{et. al.}~\cite{SPADE} incorporated VAE into their network that enables the generator to yield multi-modal images. Li~\emph{et. al.}~\cite{IMLE} adopted an implicit maximum likelihood estimation framework to alleviate the mode collapse issue of GAN, thus encouraged diverse outputs. Bansal~\emph{et. al.}~\cite{shapeandcontext} used classic tools to match the shape, context and parts from a gallery with the semantic label input in exponential ways, producing diverse results. Different from these works, we focus on semantically multi-modal image synthesis, which requires fine-grained controllability at the semantic level instead of the global level.
Gu~\emph{et. al.}~\cite{MGPE} built several auto-encoders for each face component to extract different component representations which are then merged into the next foreground generator in the task of portrait editing. Our work is highly related to this work since both methods devise to cope with the SMIS task by treating different classes with different parameters. However, our unique design of progressively decreasing the group numbers in the decoder enables our network to deal with datasets of many classes where their method is possibly incapable of. 

\vspace{0.5ex}\noindent \textbf{Group convolution.}~Previous works~\cite{AlexNet,ResNext,ShuffleNet,ShuffleNetV2,IGV2} indicate that group convolutions are advantageous for reducing the computational complexity and model parameters, thus they have been widely used in light-weight networks. Ma \emph{et. al.}~\cite{ShuffleNetV2} mentioned that excessive use of group convolutions results in large Memory Access Cost (MAC). Although it is ideal to use group convolutions with small groups or even no group convolutions in the network, we show in our experiments that completely avoiding group convolutions in the decoder is problematic for the performance on the SMIS task. Moreover, our decreasing group numbers strategy considerately alleviates the huge MAC problem so that it is applicable to real-world applications.

\section{Semantically multi-modal image synthesis}

\subsection{Problem definition.}~Let $M$ denote a semantic segmentation mask. Suppose there are $\mathcal{C}$ semantic classes in the dataset. $H$ and $W$ represent the image height and width, respectively. As a very straightforward manner of conducting label-to-image translation, the generator $G$ requires $M$ as conditional input to generate images. However, in order to support multi-modal generation, we need another input source to control generation diversity. Normally, we employ an encoder to extract a latent code $Z$ as the controller inspired by VAE~\cite{VAE}. Upon receiving these two inputs, the image output $O$ can be yield through $O=G(Z, M)$. However, in the semantically multi-modal image synthesis (SMIS) task, we aim to produce semantically diverse images by perturbing the class-specific latent code which independently controls the diversity of its corresponding class. 

\subsection{Challenge}

For the SMIS task, the key is to divide the latent code into a series of class-specific latent codes each of which controls only a specific semantic class generation. The traditional convolutional encoder is not an optimal choice because the feature representations of all classes are internally entangled inside the latent code. Even if we have class-specific latent code, it is still problematic on how to utilize the code. As we will illustrate in the experiment part, simply replacing the original latent code in SPADE \cite{SPADE} with class-specific codes has limited capability to deal with the SMIS task. This phenomenon inspires us that we need to make some architecture modifications in both the encoder and decoder to accomplish the task more effectively.

\subsection{GroupDNet}

\begin{figure*}[ht]
	\centering
	\includegraphics[width=0.85\textwidth]{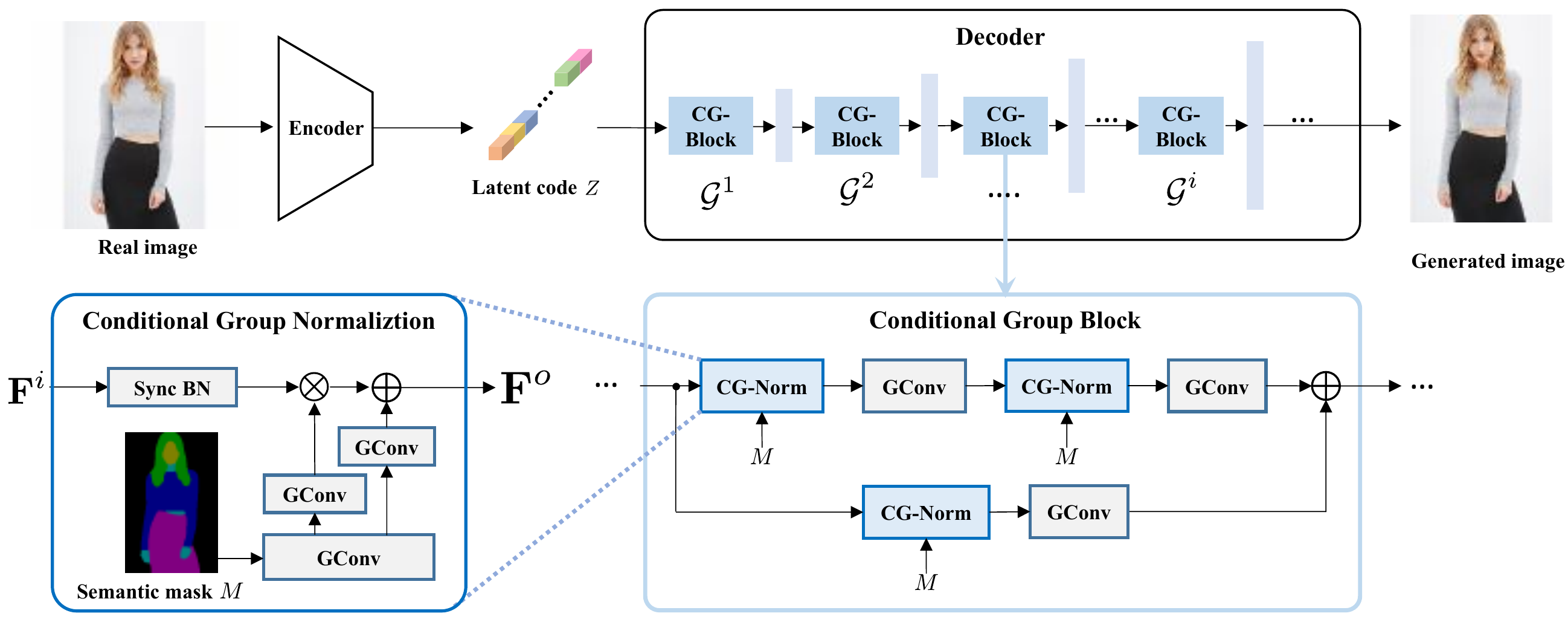}
	\caption{Architecture of our generator (GroupDNet). ``GConv'' means group convolution and ``Sync BN'' represents synchronized batch normalization. $\mathcal{G}^i$ is the group number of $i$-th layer. Note normally $\mathcal{G}^i \geq \mathcal{G}^{i+1}$ for $i\geq 1$ for GroupDNet. }
	\label{fig:model}
\end{figure*}

Based on the above analysis, we now give more details about our solution for this task---\textbf{Group Decreasing Network} (GroupDNet). The main architecture of GroupDNet takes design inspirations from SPADE~\cite{SPADE}, considering its superior performance in the label-to-image generation task. A major modification of GroupDNet is the replacement of typical convolutions to group convolutions~\cite{AlexNet} to achieve class-specific controllability.
In the following, we will first present a brief overview of the architecture of our network and then describe the modifications that we made in different components of the network.

\vspace{1ex}\noindent \textbf{Overview.}~As can be seen from Fig.~\ref{fig:model}, GroupDNet contains one encoder and one decoder. Inspired by the idea of VAE~\cite{VAE} and SPADE~\cite{SPADE}, the encoder $E$ produces a latent code $Z$ that is supposed to follow a Gaussian distribution $\mathcal{N}(0,1)$ during training. While testing, the encoder $E$ is discarded. A randomly sampled code from the Gaussian distribution substitutes for $Z$. To fulfill this, we use the re-parameterization trick~\cite{VAE} to enable a differentiable loss function during training. Specifically, the encoder predicts a mean vector and a variance vector through two fully connected layers to represent the encoded distribution. The gap between the encoded distribution and Gaussian distribution can be minimized by imposing a KL-divergence loss: 
\begin{equation}
	\label{label:klloss}
	\mathcal{L}_{\mathrm{KL}} = \mathcal{D}_{\mathrm{KL}}(E(I) || \mathcal{N} (0, 1)),
\end{equation}where $\mathcal{D}_{\mathrm{KL}}$ represents the KL divergence. 





\vspace{1ex}\noindent \textbf{Encoder.}~Let $M_c$ denote the binary mask for class $c$ and $X \in \mathbb{R}^{H\times W}$ be the input image. By splitting $X$ to different images of different semantic classes, we have
\begin{equation}
	\label{eq:split}
	X_c = M_c \cdot X.
\end{equation}
This operation reduces the dependence on $E$ to process feature disentanglement, saving more capacity to precisely encode the feature. The input to the encoder is the concatenation of these images: $S= \underset{c}{\mathrm{cat}} X_c$. 
All convolutions inside $E$ have the same number of groups, that is, the total number of classes $\mathcal{C}$. From both the input and the architecture side, we decouple different classes to be independent on each other. As a result, the encoded latent code $Z$ is comprised of the class-specific latent code $Z_c$ (a discrete part of $Z$) of all classes. And $Z_c$ serves as the controller of class $c$ in the forthcoming decoding phase.
Different from the general scheme of producing two vectors as the mean and variance prediction of the Gaussian distribution, our encoder produces a mean map and a variance map through convolutional layers to massively retain structural information in the latent code $Z$.



\vspace{1ex}\noindent \textbf{Decoder.}~Upon receiving the latent code $Z$, the decoder transforms it to natural images with the guidance of semantic labels. The question is how to leverage the semantic labels to guide the decoding phase properly. Several ways can serve this purpose, such as concatenating the semantic labels to the input or conditioning on every stage of the decoder. The former one is not suitable for our case because the decoder input has a very limited spatial size that will acutely lose many structural information of the semantic labels. We opt to the latter one and choose a typical advanced model---SPADE generator~\cite{SPADE} as the backbone of our network. As mentioned in~\cite{SPADE}, SPADE is a more general form of some conditional normalization layers~\cite{CBN,AdaIN}, and shows superior ability to produce pixel-wise guidance in semantic image synthesis. Following the general idea of using all group convolutions in the generator, we replace the convolutional layers in SPADE module with group convolutions and call this new conditional module as \emph{Conditional Group Normalization} (CG-Norm), as depicted in Fig~\ref{fig:model}. We then compose a network block called \emph{Conditional Group Block} (CG-Block) by dynamically merging CG-Norm and group convolutions. The architecture of CG-Block is also demonstrated in Fig~\ref{fig:model}.

Likewise, let $\mathbf{F}^i \in \mathbb{R}^{H^i \times W^i}$ denote the feature maps of the $i$-th layer of the decoder network and $\mathcal{G}^i$ represent the number of groups of the $i$-th layer. Moreover, $N$, $D^i$, $H^i$ and $W^i$ are the batch size, number of channels, height and width of the feature map, respectively. As demonstrated in Fig.~\ref{fig:model}, the group convolutions inside CG-Norm will transform the semantic label input to pixel-wise modulation parameters $\gamma \in \mathbb{R}^{D^i \times H^i \times W^i}$ and $\beta \in \mathbb{R}^{D^i \times H^i \times W^i}$. The feature input $\mathbf{F}^i$ will first go through a batch normalization layer \cite{batchnorm} that normalizes $\mathbf{F}^i$:
\begin{equation}
	\label{eq:bn}
	\mathrm{BN}(\mathbf{F}^i)=\gamma_{\mathrm{BN}} \left (\frac{\mathbf{F}^i-\mu(\mathbf{F}^i)}{\sigma(\mathbf{F}^i)} \right) + \beta_{\mathrm{BN}},
\end{equation} where here $\gamma_{\mathrm{BN}}, \beta_{\mathrm{BN}} \in \mathbb{R}^{D}$ are affine parameters learned from data. $\mu_d$ and $\sigma_d$ are computed across batch size and spatial dimensions for each feature channel:
\begin{equation}
	\begin{split}
		&\mu_d(\mathbf{F}^i) = \frac{1}{NH^{i}W^{i}} \sum_{n=1}^{N}\sum_{h=1}^{H^i}\sum_{w=1}^{W^i} \mathbf{F}^i_{ndhw}\\
		&\sigma_d(\mathbf{F}^i) = \sqrt{\frac{1}{NH^{i}W^{i}} \sum_{n=1}^{N}\sum_{h=1}^{H^i}\sum_{w=1}^{W^i}  (\mathbf{F}^i_{ndhw})^2 - (\mu_d(\mathbf{F}^i))^2}
	\end{split}
\end{equation}
Afterwards, the output $\mathrm{BN}(\mathbf{F}^i)$ interacts with previously predicted pixel-wise $\gamma$ and $\beta$, yielding a new feature map $\mathbf{F}^o$ with semantic information inserted. Taking Eq.~\ref{eq:bn} into account, 
\begin{equation}
	\begin{split}
		\mathbf{F}^o &=\gamma \cdot \mathrm{BN}(\mathbf{F}^i)+\beta \\
		&=\gamma \cdot \gamma_{\mathrm{BN}} \left (\frac{\mathbf{F}^i-\mu(\mathbf{F}^i)}{\sigma(\mathbf{F}^i)} \right) + (\gamma \cdot \beta_{\mathrm{BN}} + \beta).
	\end{split}
\end{equation}

When $i$ becomes larger, the group number is finally reduced to 1. After a regular convolution, the feature is mapped to a three-channel RGB image $O$. 

\subsection{Other solutions}

\begin{figure*}[ht]
	\centering
	\includegraphics[width=0.9\textwidth]{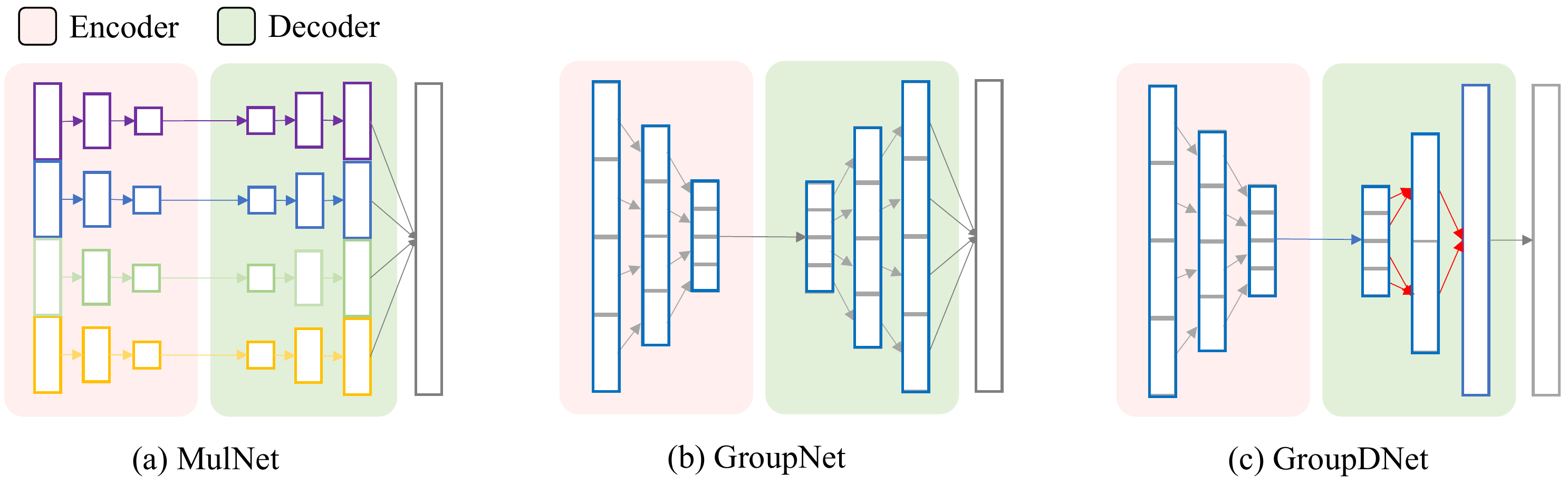}
	\caption{An illustration of MulNet (a), GroupNet (b) and GroupDNet (c). Note the last layers of MulNet and GroupNet are fusing modules that are comprised of several normal convolutional layers to fuse results of different classes.}
	\label{fig:groupdnet}
\end{figure*}

\label{sec:other_solutions}

Aside from GroupDNet, a simple solution to perform the SMIS task is to build a set of encoders and decoders, each of which focus on a specific semantic class, as demonstrated in Fig.~\ref{fig:groupdnet} (a). The underlying idea is to treat each class independently and then fuse the results of different sub-networks. For simplicity, we call such networks as \emph{Multiple Networks} (MulNet).
Another alternative with a similar idea is to use group convolution~\cite{AlexNet} throughout the network. As depicted in Fig.~\ref{fig:groupdnet} (b), replacing all the convolutions in the encoder and the decoder with group convolutions~\cite{AlexNet} and setting the group number equal to the class number present the~\emph{Group Network} (GroupNet). It is theoretically equivalent to MulNet if the channel number in every group is equal to those of the corresponding layer in a single network of MulNet. Fig.~\ref{fig:groupdnet} (c) illustrates our GroupDNet. The primary difference between GroupDNet and GroupNet is the monotonically decreasing number of groups in the decoder. 
Although this modification seems to be simple, it brings several noticeable benefits, mainly in the following three aspects:

\vspace{1ex}\noindent \textbf{Class balance.}~It is worth noticing that different classes have a different number of instances~\cite{DeepFashion,Cityscapes,ADE20K} and require different network capacity to model these classes. It is difficult for MulNet and GroupNet to find a suitable network design to balance all the classes. More importantly, not all the classes appear in one image. In this case, MulNet and GroupNet inevitably waste a lot of computational resources because they have to activate all the sub-networks or sub-groups for all the classes during training or testing. However, in GroupDNet, unbalanced classes share parameters with their neighbor classes, hugely alleviating the class imbalance problem.

\vspace{1ex}\noindent \textbf{Class correlation.}~In natural worlds, a semantic class usually has relationships with other classes, \eg, the color of grass and the color of tree leaves are similar, and buildings influence the sunshine on the roads in their vicinity,~\etc.
To generate plausible results, both of MulNet and GroupNet have a fusion module (several regular convolutions in our case) at the end of the decoder to merge features of different classes into one image output. In general, the fusion module roughly considers the correlations of different classes. However, we argue it is not sufficient because the correlation of different classes is too complex to be fully explored by using such a simple component with restricted receptive fields. An alternative is to use some network modules like self-attention block to capture long-range dependencies of the image, but its prohibitive computation hinders its usage in such scenarios~\cite{SAGAN}. GroupDNet, however, carves these relationships throughout the decoder; hence, it exploits the correlations more accurately and thoroughly. As a result, the generated images of GroupDNet are better and more realistic than those generated by the other two methods.


\vspace{1ex}\noindent \textbf{GPU memory.}~In order to guarantee that every single network of MulNet, or the grouped parameters for each class in GroupNet have sufficient capacity, the channel numbers in total will increase significantly with the increase of class number. Up to a limit, the maximum GPU memory of a graphics card would no longer be able to hold even one sample. As we roughly estimate on the ADE20K dataset \cite{ADE20K}, one Tesla V100 graphics card cannot hold the model with sufficient capacity even when batch size is set to 1. However, the problem is less severe in GroupDNet because different classes share parameters, thus it is unnecessary to set so many channels for each class.





\subsection{Loss function}
We adopt the same loss function as SPADE~\cite{SPADE}:
\begin{equation}
\mathcal{L}_{\mathrm{full}}=\arg\underset{G}{\min}\, \underset{D}{\max}\, \mathcal{L}_{\mathrm{GAN}} + \lambda_1\mathcal{L}_{\mathrm{FM}} + \lambda_2 \mathcal{L}_{\mathrm{P}} + \lambda_3 \mathcal{L}_{\mathrm{KL}}.
\end{equation}
The $\mathcal{L}_{\mathrm{GAN}}$ is the hinge version of GAN loss, and $\mathcal{L}_{\mathrm{FM}}$ is the feature matching loss between the real and synthesized images. Specifically, we use a multiple-layer discriminator to extract features from real and synthesized images. Then, we calculate the $L_1$  distance between these paired features. Likewise, $\mathcal{L}_{\mathrm{P}}$ is the perceptual loss proposed for style transfer~\cite{perceptualloss}. A pre-trained VGG network~\cite{VGG} is used to get paired intermediate feature maps, and then we calculate the $L_1$ distance between these paired maps. $\mathcal{L}_{\mathrm{KL}}$ is the KL-divergence loss term as Eq.~\ref{label:klloss}. We set $\lambda_1=10, \lambda_2=10, \lambda_3= 0.05$, the same as SPADE~\cite{SPADE}.

\section{Experiments}


\subsection{Implementation details }
We apply Spectral Normalization \cite{SpectralNorm} to all the layers in both the generator and discriminator. The learning rates for the generator and discriminator are set to 0.0001 and 0.0004, respectively, following the two time-scale update rule \cite{TTUR}. We use the Adam optimizer \cite{Adam} and set $\beta_1 = 0, \beta_2 = 0.9$. All the experiments are conducted on at least 4 P40 GPUs. Besides, we use synchronized batch normalization to synchronize the mean and variance statistics across multiple GPUs. More details, such as the detailed network design and more hyper-parameters, are given in the supplementary materials.

\subsection{Datasets}

We conduct experiments on three very challenging datasets, including DeepFashion \cite{DeepFashion}, Cityscapes \cite{Cityscapes}, and ADE20K \cite{ADE20K}. We choose DeepFashion because this dataset shows lots of diversities among all semantic classes, which is naturally suitable for assessing the model's ability to conduct multi-modal synthesis. Consequently, we compare with several baseline models on this dataset to evaluate the superior power of our model on the SMIS task. The size of the images in Cityscapes are quite large, so it is proper to test the model's ability to produce high-resolution images on this dataset. ADE20K is extremely challenging for its massive number of classes, and we find it hard to train MulNet and GroupNet on ADE20K with our limited GPUs.
More details can be found in the supplementary materials.




\subsection{Metrics }
\vspace{0.5ex}\noindent \textbf{Mean SMIS Diversity.}~In order to evaluate the performance of a model designed for the SMIS task, we introduce two new metrics named: \emph{mean Class-Specific Diversity} (mCSD) and \emph{mean Other-Classes Diversity} (mOCD). We design the new metrics based on the LPIPS metric~\cite{LPIPS}, which is used to assess the generation diversity of a model by computing the weighted $\mathcal{L}_2$ distance between deep features of image pairs. For the same semantic label input, we generate $n$ images for each semantic class by only modulating the latent code $Z_c$ for the semantic class $c$. Therefore, we have a set of images $S=\{I_1^1, ... , I_1^n, ..., I_\mathcal{C}^1, ..., I_\mathcal{C}^n \}$. 
Finally, mCSD and mOCD are calculated by
\begin{equation}
\begin{aligned}
\mathrm{mCSD} = \frac{1}{\mathcal{C}}\sum_{c=1}^{\mathcal{C}} L_c, \ \mathrm{mOCD} = \frac{1}{\mathcal{C}}\sum_{c=1}^{\mathcal{C}} L_{\ne c}. 
\end{aligned}
\end{equation}
where $L_c$ is the average LPIPS distance~\cite{LPIPS} of the semantic area of class $c$ between sampled $m$ pairs and $L_{\ne c}$ represents the average LPIPS distance~\cite{LPIPS} in the areas of all other classes between the same pairs. 
In our settings, we set $n = 100, m = 19$ following \cite{BiCycleGAN,MUNIT}. ImageNet pre-trained AlexNet~\cite{AlexNet} is served as the deep feature extractor. Higher performance on the SMIS task demands a high diversity of the specific semantic areas (high mCSD) as well as a low diversity of all other areas (low mOCD). Besides, we also report the overall LPIPS distance by producing globally diverse results for the same semantic labels.

\vspace{1ex}\noindent \textbf{Human Evaluation Metrics.}~We further introduce human evaluation to evaluate whether the generative model performs well in the SMIS task. We recruit 20 volunteers that have research experience in generation tasks. We show them an input mask along with two images generated from only one model. The two images are multi-modal results which only vary in the areas of one random semantic class. The volunteers judge whether the given two images only vary in one semantic class. The percentage of pairs that are judged to be semantically different in only one semantic class represents the human evaluation of a model's performance on the SMIS task. We abbreviate this metric as \textbf{SHE} (\textbf{S}MIS \textbf{H}uman \textbf{E}valuation). For each phase, the volunteers are given 50 questions of unlimited answering time. 



\vspace{1ex}\noindent \textbf{Fr\'{e}chet Inception Distance.}~
We use Fr\'{e}chet Inception Distance (FID) \cite{FID} to calculate the distance between the distributions of synthesized results and the distribution of real images. Lower FID generally hints better fidelity of the generated images.

\vspace{1ex}\noindent \textbf{Segmentation Performance.}~It is reasonable that the predicted labels of the generated images are highly similar to those of the original images if they look realistic. Therefore, we adopt the evaluation protocol from previous work \cite{PhotographicImageSynthesis,pix2pixHD,SPADE} to measure the segmentation accuracy of the generated images. We report results on the mean Intersection-over-Union (mIoU) and pixel accuracy (Acc) metrics without considering the classes that can be ignored. Images are evaluated using well-trained segmentation models UperNet101 \cite{UperNet} for ADE20K,  DRN-D-105 \cite{DRN-D} for Cityscapes, off-the-shelf human parser CIHP~\cite{CIHP} for DeepFashion.

\subsection{Results}

Besides the following sections, we have more justification of our model design in the supplementary materials for the reference of interested readers.

\subsubsection{Comparison on SMIS}

\begin{figure}[t]
	\centering
	\includegraphics[width=0.47\textwidth]{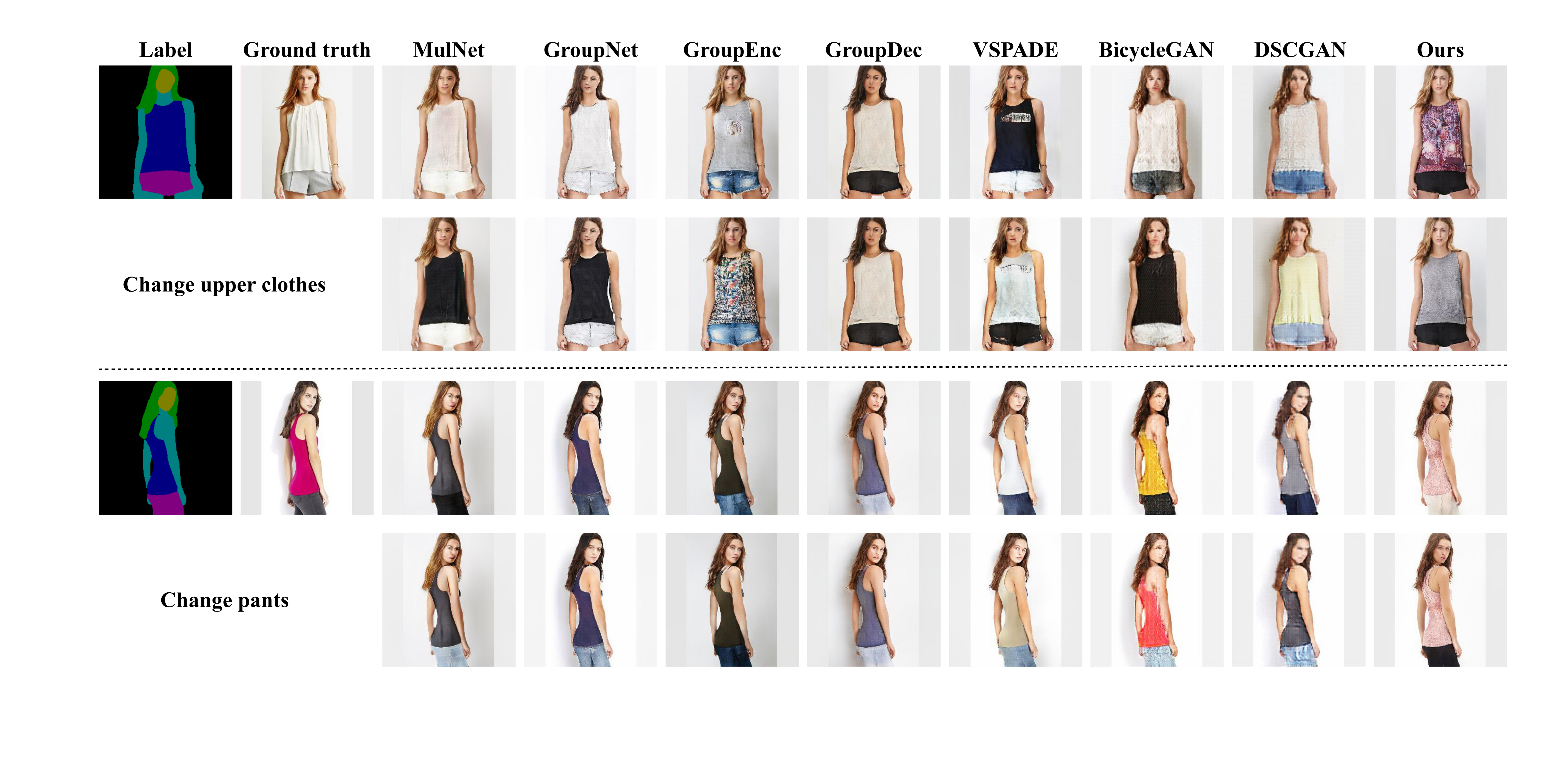}
	\caption{Qualitative comparison between GroupDNet and other baseline models. The first two rows represent the results of different models by changing their upper-clothes latent code while the last two rows represent their results of changing the pants latent code. Note, for those models which have no class-specific controller such as VSPADE, we alter their overall latent codes to generate different images.}
	\label{fig:ablation}
\end{figure}

A basic requirement for models that potentially could be modified for the SMIS task is that they should possess the ability to conduct multi-modal image synthesis. We compare with several methods that support multi-modal image synthesis to demonstrate the superiority of GroupDNet:

\begin{itemize}
	\item[-]\textbf{Variational SPADE~\cite{SPADE}} (\textbf{VSPADE})~has an image encoder processing a real image to a mean and a variance vector where a KL divergence loss is applied to support multi-modal image synthesis. Detailed description can be found in their paper; 
	\item[-] \textbf{BicycleGAN~\cite{BiCycleGAN}}~maps the given image input into a latent code, which is later combined with a label input to produce outputs. Since the latent code is constrained by a KL divergence loss, it could be substituted by a random sample from the Gaussian distribution; 
	\item[-] \textbf{DSCGAN~\cite{dscgan}}~is complementary to BicycleGAN by introducing an explicit regularization upon the generator, trying to alleviate the mode collapse issue of previous models.
\end{itemize}

\noindent Aside from MulNet and GroupNet described in Sec.~\ref{sec:other_solutions}, we also conduct two further experiments by replacing the convolutions in the encoder/decoder of the VSPADE model to group convolutions with group numbers set equal to the dataset class number, denoted as \textbf{GroupEnc/GroupDec}, respectively. Note MulNet, GroupNet, GroupEnc, GroupDec and VSPADE are trained with the same kind of multi-scale discriminator~\cite{pix2pixHD} and training settings as GroupDNet. To fairly compare the performances, we balance the number of  parameters of these models to mitigate the suspicion that the performance improvements are brought by using more parameters. For BicycleGAN and DSCGAN, we adopt their original training and testing protocols. 

Quantitative and qualitative results are given in Tab.~\ref{tab:ablation_baseline} and Fig.~\ref{fig:ablation}, respectively. The quantitative results demonstrate the overall superiority of GroupDNet. Generally, GroupDNet exhibits the best image quality (lowest FID) and overall diversity (highest LPIPS). In terms of the performance on the SMIS task, MulNet and GroupNet are slightly better than GroupDNet, given evidence that they have either larger mCSD or lower mOCD. However,  the image quality of MulNet and GroupNet is not satisfactory (high FID) and MulNet shows much lower FPS than GroupDNet. In terms of the SHE metric, GroupDNet is also very competitive to MulNet and GroupNet. 
Although VSPADE, has rather large mCSD, its mOCD is also very large, indicating that it performs unsatisfactorily on the SMIS task. The same phenomenon is also observed in BicycleGAN and DSCGAN and their FIDs are relatively much higher than VSPADE, showing the advantage of the SPADE architecture. From the high mOCD values of VSPADE and GroupDec, whose encoders are composed from regular convolutions, we conclude that group encoder serves as a key to the high performance of the SMIS task. However, the exceptional performance of GroupDNet suggests that the group decreasing modification in the decoder is also effective and brings further performance boost when compared to GroupEnc.
Gathering these information, GroupDNet is a good trade-off model considering the speed, visual quality and the performance on the SMIS task. 

According to the qualitative results, it is obvious that MulNet, GroupNet, GroupEnc and GroupDNet are able to generate semantically multi-modal images while others cannot. However, the image quality of MulNet, GroupNet, BicycleGAN and DSCGAN is far from satisfaction because their images are visually implausible. GroupEnc is better in image quality but it degrades in the SMIS task. It can be seen from the first two rows in Fig.~\ref{fig:ablation} that, when the upper clothes are changed to another style, GroupEnc slightly changes the color of the short jeans pants as well. 

\begin{table}[tb]
	\centering
	\resizebox{0.48\textwidth}{!}{
		\begin{tabular}{|l|c|c|c|c|c|c|c|} 
			\hline
			Models & FID$\downarrow$  & mCSD$\uparrow$ & mOCD$\downarrow$ & LPIPS$\uparrow$ & SHE$\uparrow$ & Speed$\uparrow$ & \# Param$\downarrow$ \\ \hline
			MulNet & 12.07 & 0.0244 & 0.0019 & 0.202 & 79.2 & 6.3 & 105.1 \\
			GroupNet & 12.58 & 0.0276 & \textbf{0.0017} & 0.203 & \textbf{83.7} & 8.2 & 97.7 \\
			Group Enc & 10.83 & 0.0232 & 0.0065 & 0.217& 69.3 & 19.6 & 105.5 \\ 
			Group Dec & 9.84 & 0.0003 & 0.0257 & 0.206 & 26.4 & 12.1 & 111.3  \\ 
			VSPADE~\cite{SPADE} & 10.02 & 0.0304 & 0.1843 &0.207 & 23.6 & 20.4 & 106.8 \\
			BicycleGAN~\cite{BiCycleGAN}& 40.07 & \textbf{0.0316} & 0.2147 & \textbf{0.228} & 24.8 & 66.9 & \textbf{58.4}   \\ 
			DSCGAN~\cite{dscgan} & 38.40 & 0.0245 & 0.1560 & 0.163 & 27.6  &   \textbf{67.2} & \textbf{58.4} \\ \hline
			\textbf{GroupDNet} & \textbf{9.50} & 0.0264 & 0.0033 & \textbf{0.228} & 81.2 & 12.2 & 109.1 \\ \hline
		\end{tabular}
	}
	\caption{Quantitative comparison results with baseline models. ``SHE'' means human evaluation of a model's performance on SMIS task. We use Frame Per Second (FPS) to represent the ``Speed'' of the model. ``\# Param'' means the number of parameters, whose unit is ``M'', denoting million. For mCSD, the higher, the better. For mOCD, the lower, the better.}
	\label{tab:ablation_baseline}
\end{table}

\subsubsection{Comparison on label-to-image translation}
\begin{figure}[t]
	\centering
	\includegraphics[width=0.48\textwidth]{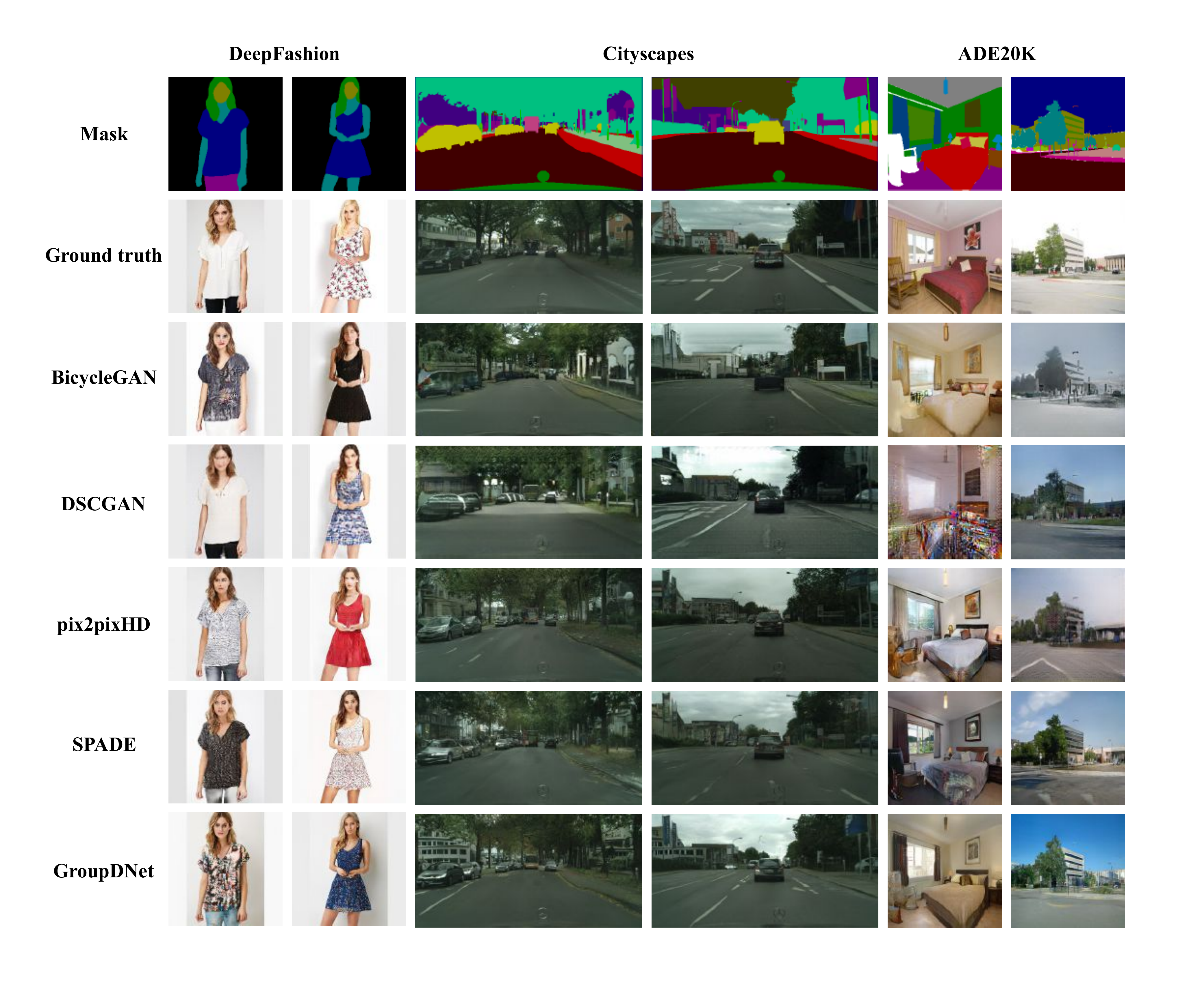}
	\caption{Qualitative comparison with the SOTA label-to-image methods. From top to bottom, the images represent the experiments on DeepFashion, Cityscapes and ADE20K, respectively.}
	\label{fig:comparisons}
\end{figure}

\begin{figure*}[h]
	\vspace{-2cm}
	\centering
	\includegraphics[width=\textwidth]{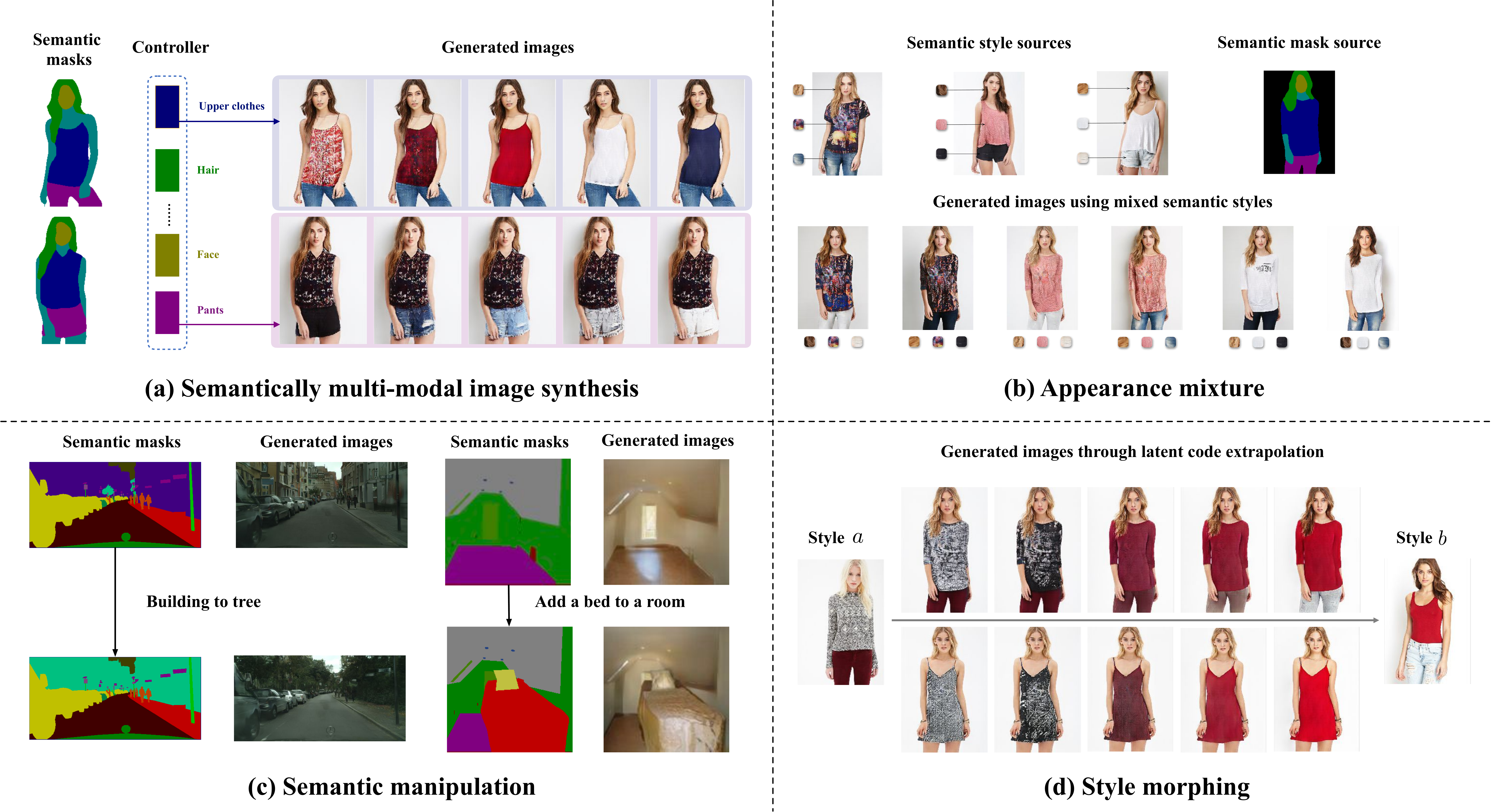}
	\caption{ Exemplar applications of the proposed method. (a) Demonstration of the semantically multi-modal image synthesis (SMIS) task. (b) Application of our SMIS model in appearance mixture. Our model extracts styles of different semantic classes from different sources and generates a mixed image by combining these semantic styles with the given semantic mask. (c) Application of our SMIS model in semantic manipulation. (d) Application of our SMIS model in image extrapolation. \textbf{Zoom in} for better details.
		\label{fig:applications}
	}
\end{figure*}

In this section, we mainly assess the generated image quality of our method by comparing with some label-to-image methods on the FID, mIoU and Accuracy metrics. We choose four very recent state-of-the-art methods: BicycleGAN~\cite{BiCycleGAN}, DSCGAN~\cite{dscgan}, pix2pixHD~\cite{pix2pixHD} and SPADE~\cite{SPADE}, as the comparison methods. Comparisons are performed across the DeepFashion, Cityscapes and ADE20K datasets. We evaluate the performance of their well-trained models downloaded from their official GitHub repositories if they have. 
For those experiments not included in their original papers, we follow their codes and run the experiments with similar settings to GroupDNet. 
Quantitative results are shown in Tab.~\ref{tab:comparison}. In general, as our network is built based on SPADE, it maintains nearly the same performance as SPADE on the DeepFashion and Cityscapes datasets. While on the ADE20K dataset, our method is inferior to SPADE but still outperform other methods. This phenomenon on the one hand shows the superiority of the SPADE architecture and on the other hand also exposes even GroupDNet still struggles to handle datasets with a huge number of semantic classes.


Qualitative comparisons on DeepFashion, Cityscapes and ADE20K are shown in Fig.~\ref{fig:comparisons}. In general, the images generated by GroupDNet are more realistic and plausible than others. 
These visual results consistently show the high image quality of GroupDNet's generated images, verifying its efficacy on various datasets.

\begin{table}[tb]
	\centering
	\resizebox{0.48\textwidth}{!}{
		\begin{tabular}{|l|c|c|c|c|c|c|c|c|c|}
			\hline
			\multirow{2}{*}{Method}&
			\multicolumn{3}{c|}{DeepFashion}&
			\multicolumn{3}{c|}{Cityscapes}&
			\multicolumn{3}{c|}{ADE20K} \\
			\cline{2-4} \cline{5-7} \cline{8-10} 
			& mIoU$\uparrow$ & Acc$\uparrow$ & FID$\downarrow$ & mIoU$\uparrow$ & Acc$\uparrow$ & FID$\downarrow$ & mIoU$\uparrow$ & Acc$\uparrow$ & FID$\downarrow$ \\
			\hline
			BicycleGAN~\cite{BiCycleGAN} & 76.8 & 97.8 & 40.07 & 23.3 & 75.4 & 87.74 & 4.78 & 29.6 & 87.85  \\
			DSCGAN~\cite{dscgan} & 81.0 & 98.3 & 38.40 & 37.8 & 86.7 & 67.77 & 10.2 & 58.8 & 83.98  \\
			pix2pixHD~\cite{pix2pixHD} & 85.2 & 98.8 & 17.76 & 58.3  & 92.5 &  78.24 & 27.6  & 75.7  & 55.9    \\
			SPADE~\cite{SPADE} & 87.1 & \textbf{98.9} & 10.02 & \textbf{62.3}  & 93.5 & 58.10 &  \textbf{42.0} & \textbf{81.4} & \textbf{33.49}  \\  \hline
			\textbf{GroupDNet} & \textbf{87.3} & \textbf{98.9} & \textbf{9.50} & \textbf{62.3} & \textbf{93.7} & \textbf{49.81} & 30.4 & 77.1 & 42.17 \\ \hline
		\end{tabular}
	}
	\caption{Quantitative comparison with label-to-image models. The numbers of pix2pixHD and SPADE are collected by running the evaluation on our machine instead of their papers.}
	\label{tab:comparison}
\end{table}

\subsubsection{Applications}


Since GroupDNet contributes more user controllability to the generation process, it can also be applied to lots of exciting applications in addition to the SMIS task, which are demonstrated as follows. More results are available in the supplementary materials.

\vspace{0.5ex}\noindent \textbf{Appearance mixture.}~By utilizing the encoder in GroupDNet during inference, we can gather the distinct styles of a person's different body parts. Every combination of these styles presents a distinct person image, given a human parsing mask. In this way, we can create thousands of diverse and realistic person images given a person image gallery. This application is demonstrated in Fig.~\ref{fig:applications}(b). A demo video can be found in our code repository. 

\vspace{0.5ex}\noindent \textbf{Semantic manipulation.}~Similar to most label-to-image methods~\cite{pix2pixHD,SPADE,IMLE}, our network also supports semantic manipulation. As exemplified in Fig.~\ref{fig:applications} (c), we can insert a bed in the room or replace the building with trees, \etc.

\vspace{0.5ex}\noindent \textbf{Style morphing.}~Feeding two real images to the encoder generates two style codes of these images. By extrapolating between these two codes, we can generate a sequence of images that progressively vary from image $a$ to image $b$, depicted in Fig.~\ref{fig:applications} (d).



\section{Conclusion and future work}

In this paper, we propose a novel network for semantically multi-modal synthesis task, called GroupDNet. Our network unconventionally adopts all group convolutions and modifies the group numbers of the convolutions to decrease in the decoder, considerably improving the training efficiency over other possible solutions like multiple generators.

Although GroupDNet performs well on semantically multi-modal synthesis task and generates results with relatively high quality, there are still some problems remained to be solved. First, it requires more computational resources to train compared to pix2pixHD and SPADE though it is nearly 2 times faster than multiple generators networks. Second, GroupDNet still has difficulty in modeling different layouts of a specific semantic class for datasets with limited diversity, even though it demonstrates some low-level variations like illumination, color, and texture, \etc. 

\section*{Acknowledgment}
We thank Taesung Park for his kind help to this project. This work was supported by NSFC 61573160, to Dr. Xiang Bai by the National Program for Support of Top-notch Young Professionals and the Program for HUST Academic Frontier Youth Team.

{\small
\bibliographystyle{ieee_fullname}
\bibliography{egbib}
}

\clearpage
 \begin{center}
	\Large \textbf{Appendix}
\end{center}
\appendix

\section{Implementation details}
\begin{figure}[tb]
	\centering
	\includegraphics[width=0.4\textwidth]{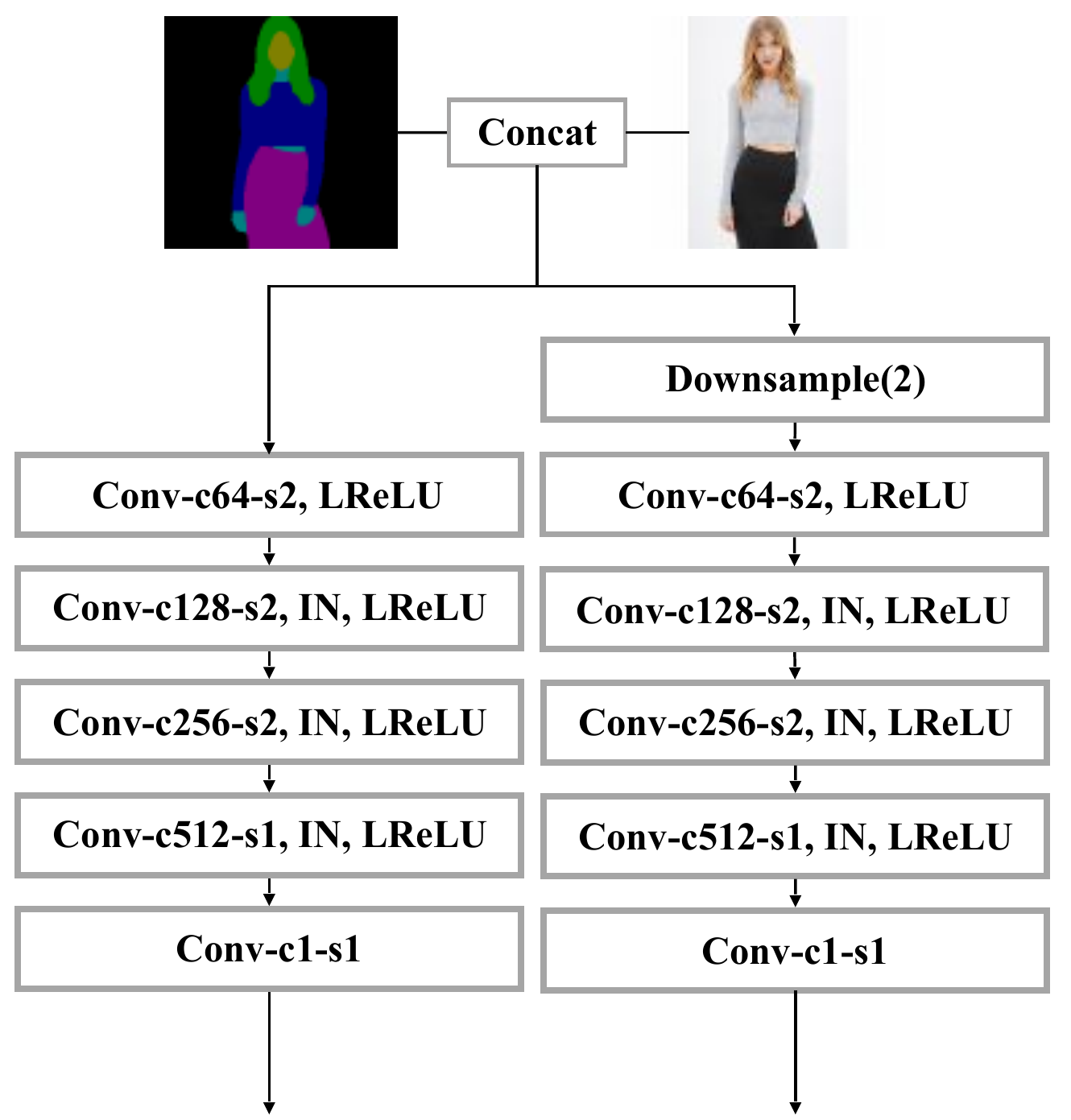}
	\caption{Architecture of our discriminator for all three datasets. Note ``Conv'' means convolutional layer. The numbers after ``-c'', ``-s'' and ``-g'' represent the channel number, stride and group number of the corresponding convolution. If not specified, the default kernel size, stride, padding and group number of the convolutional layer are 4, 2, 2, 1, respectively. ``IN'' represents instance normalization layer and ``LReLU'' means leaky ReLU layer. ``Downsample($\cdot$)'' means an average pooling layer with kernel size set to the number inside the bracket. }
	\label{fig:discriminator_arc}
\end{figure}

\begin{figure}[tb]
	\centering
	\includegraphics[width=0.3\textwidth]{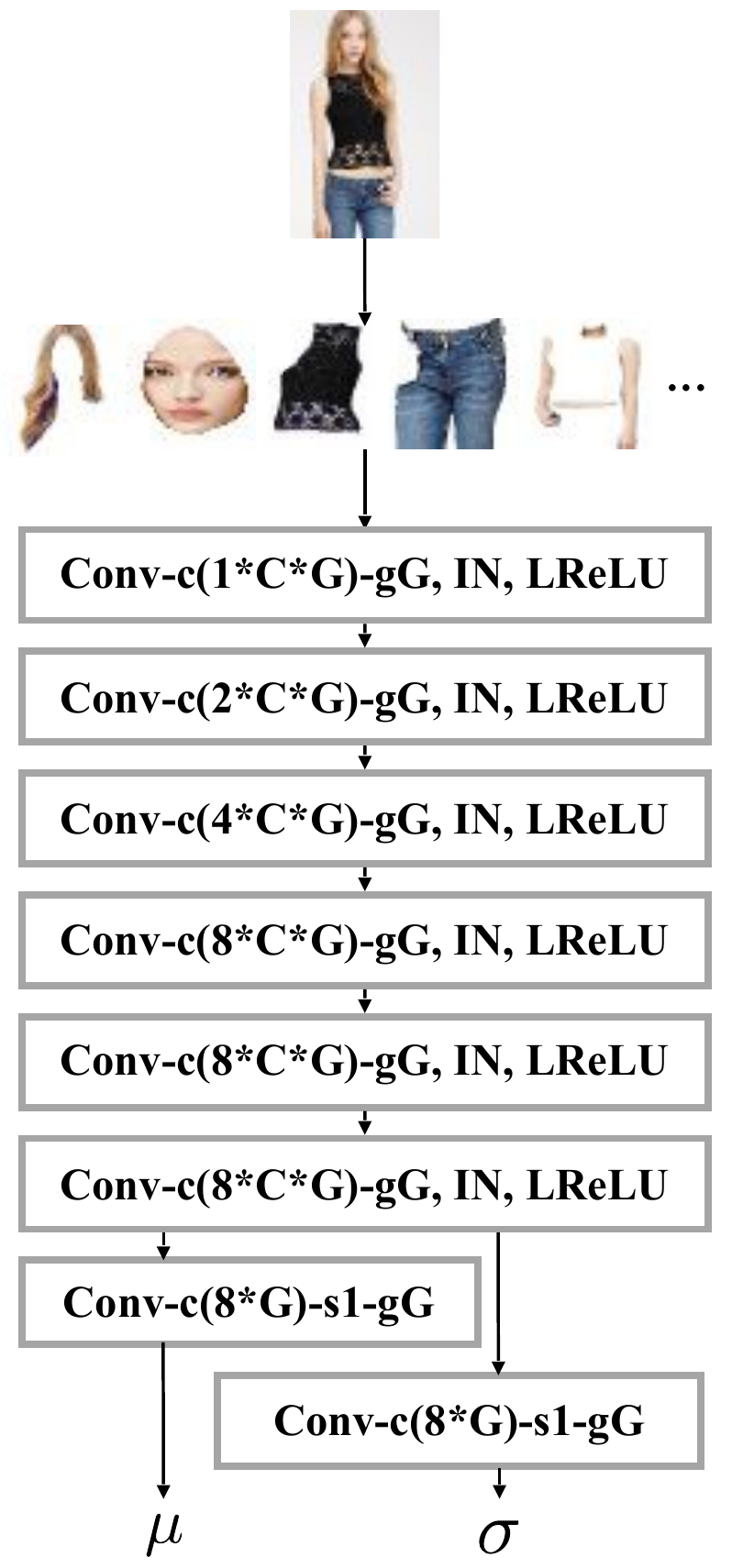}
	\caption{Architecture of our encoder for three datasets. Note ``Conv'' means convolutional layer. The numbers after ``-c'', ``-s'' and ``-g'' represent the channel number, stride and group number of the corresponding convolution. If not specified, the default kernel size, stride, padding and group number of the convolutional layer are 3, 2, 1, 1, respectively. ``IN'' represents instance normalization layer and ``LReLU'' means leaky ReLU layer. Here ``C'' and ``G'' are pre-defined numbers for each datasets, which are given in Tab.~\ref{tab:defined_numbers}.}
	\label{fig:encoder_arc}
\end{figure}

\begin{figure}[tb]
	\centering
	\includegraphics[width=0.4\textwidth]{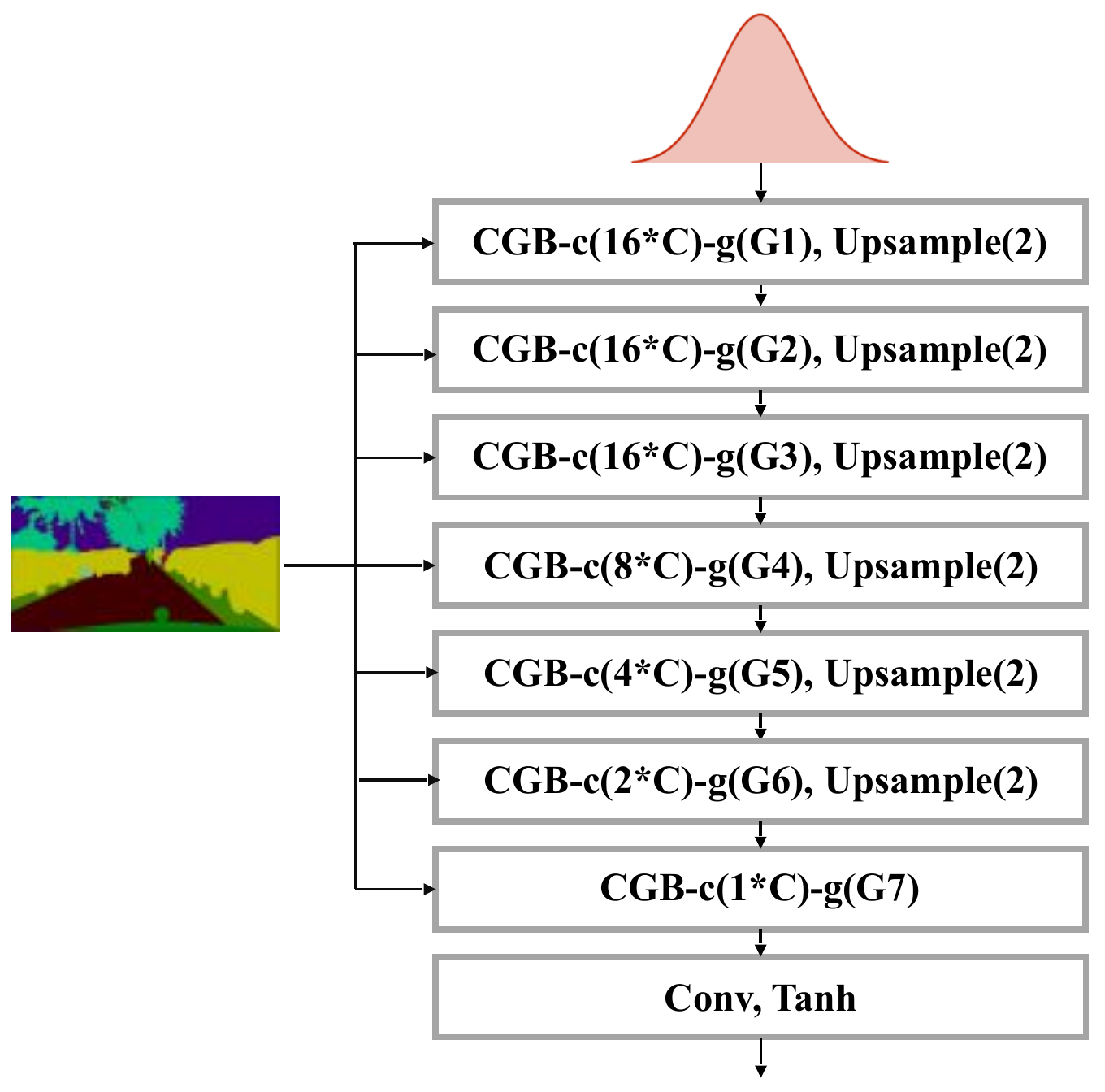}
	\caption{Architecture of our decoder for the DeepFashion and Cityscapes datasets. Note ``CGB'' means our conditional group block (CG-Block). The numbers after ``-c'', ``-s'' and ``-g'' represent the channel number, stride and group number of the corresponding convolution. If not specified, the default kernel size, stride, padding and group number of the convolutional layer are 3, 1, 1, 1, respectively. After each CG-Norm inside CG-Block, there follows a ReLu layer. ``Upsample($\cdot$)'' means a nearest neighbor upsampling layer with kernel size set to the number inside the bracket. Here ``C'' and ``G'' are pre-defined numbers for each datasets, which are given in Tab.~\ref{tab:defined_numbers}.}
	\label{fig:decoder_fashioncity}
\end{figure}

\begin{figure}[tb]
	\centering
	\includegraphics[width=0.4\textwidth]{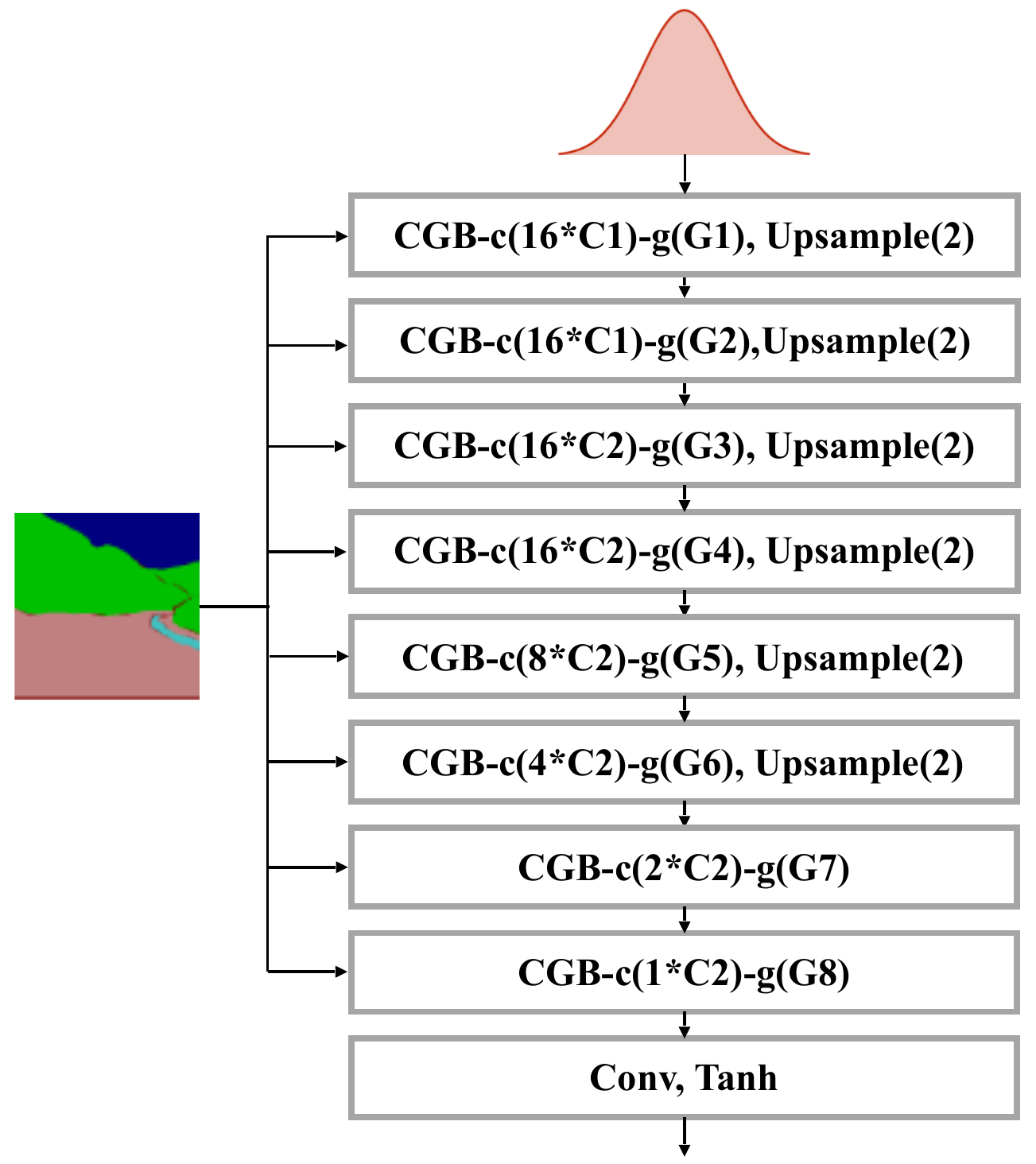}
	\caption{Architecture of our decoder for the ADE20K datasets. Note ``CGB'' means our conditional group block (CG-Block). The numbers after ``-c'', ``-s'' and ``-g'' represent the channel number, stride and group number of the corresponding convolution. If not specified, the default kernel size, stride, padding and group number of the convolutional layer are 3, 1, 1, 1, respectively. After each CG-Norm inside CG-Block, there follows a ReLu layer. ``Upsample($\cdot$)'' means a nearest neighbor upsampling layer with kernel size set to the number inside the bracket. Here ``C'' and ``G'' are pre-defined numbers for each datasets, which are given in Tab.~\ref{tab:defined_numbers}.}
	\label{fig:decoder_ade20k}
\end{figure}

\vspace{1ex} \noindent \textbf{Network architectures.}~In this section, we give detailed network designs for each dataset. 
We demonstrate the architecture of the discriminator in Fig.~\ref{fig:discriminator_arc}. Note the architecture of the discriminator holds the same for different datasets.
In Fig.~\ref{fig:encoder_arc}, we demonstrate the encoder architecture for different datasets. 
Fig.~\ref{fig:decoder_fashioncity} depicts the architectures of the decoders for the DeepFashion and Cityscapes, while Fig.~\ref{fig:decoder_ade20k} shows the architecture of the decoder for ADE20K. Since ADE20K has so many classes, we bring down the channel number for each group to avoid massive GPU usage. In this case, the overall network capacity decreases, and we assume it's not helpful to the results. Therefore, we add some additional convolutional layers to enlarge the network capacity; thus, this makes the architecture of the decoder for ADE20K is different from those of the other two datasets. 

\vspace{1ex} \noindent \textbf{Training details.}~We train all experiments on DeepFashion for 100 epochs, where in the first 60 epochs, the learning rates for both the generator and discriminator maintain the same while linearly decay to 0 in the last 40 epochs. For Cityscapes and ADE20K datasets, we follow the training settings of SPADE~\cite{SPADE} to train 200 epochs, where the learning rates linearly decay to 0 from 100 to 200 epochs. The image sizes are $256 \times 256$, except the Cityscapes at $512 \times 256$. The batch size for DeepFashion and Cityscapes is 32 while 16 for ADE20K due to the large number of channels to meet the requirements of sufficient capacity for the 150 classes. The network weights are initialized with Glorot initialization~\cite{Glorot} following SPADE~\cite{SPADE}.

\vspace{1ex} \noindent \textbf{Group number selection strategy.}~Actually, it is hard to devise a programmatic strategy to decide the decreasing numbers, under restrictions of the capacity of GPU memory, batch size, and the number of parameters \etc. However, we still followed two rules to design the group numbers: 1) the numbers decrease drastically in the first several layers of the decoder to largely reduce the computational cost; 2) the group number in the previous layer is either equal or 2 times of that in the next layer.

\section{Datasets}


\vspace{1ex}\noindent \textbf{DeepFashion~\cite{DeepFashion}.}~DeepFashion (In-shop Clothes Retrieval Benchmark) contains of 52,712 person images with fashion clothes. We select about 29,000 training and 2,500 validation images. After that, we use an off-the-shelf human parser~\cite{CIHP} pre-trained on the LIP dataset~\cite{LIP} to get segmentation maps. Specifically, given an input image, we first get its segmentation map, then re-organize the map into eight categories: hair, face, skin (including hands and legs), top-clothes, bottom-clothes, socks, shoes, and background. At the same time, we filter out the images with some rare attributes like a glove, and so on. We choose DeepFashion because this dataset shows lots of diversities of all semantic classes, which is naturally suitable for assessing the model's ability to conduct multi-modal synthesis.

\vspace{1ex}\noindent \textbf{Cityscapes~\cite{Cityscapes}.}~Cityscapes dataset~\cite{Cityscapes} has 3,000 training images and 500 validation images, collected from German cities. The size of the images in Cityscapes are quite large, so it is proper to test the model's ability to produce high-resolution images on this dataset.

\vspace{1ex}\noindent \textbf{ADE20K~\cite{ADE20K}.}~ADE20K dataset~\cite{ADE20K} contains 20,210 training and 2000 validation images. This dataset is extremely challenging for many tasks because it contains up to 150 semantic classes. ADE20K is extremely challenging for its massive number of classes, and we find it hard to train MulNet and GroupNet on ADE20K with our limited GPUs.


\section{Additional results}
In Fig~\ref{fig:ablation}, we show more ablation qualitative results on DeepFashion. The conclusions are basically the same as we put in the main submission. One thing to note is that compared to MulNet, GroupNet, and GroupEnc, our GroupDNet has better color, style, and illumination consistencies due to its design consideration for carving the correlation among different classes. Likewise, GroupDec and VSAPDE seem to have the ability to consider class correlations just as GroupDNet, because the regular convolutions in their decoders help to discover the relationships. But they instead lose strong SMIS controllability, unlike GroupDNet.
These results firmly verify the efficacy of GroupDNet and show its balanced trade-off between SMIS controllability and image quality.

\begin{figure*}[tb]
	\centering
	\includegraphics[width=\textwidth]{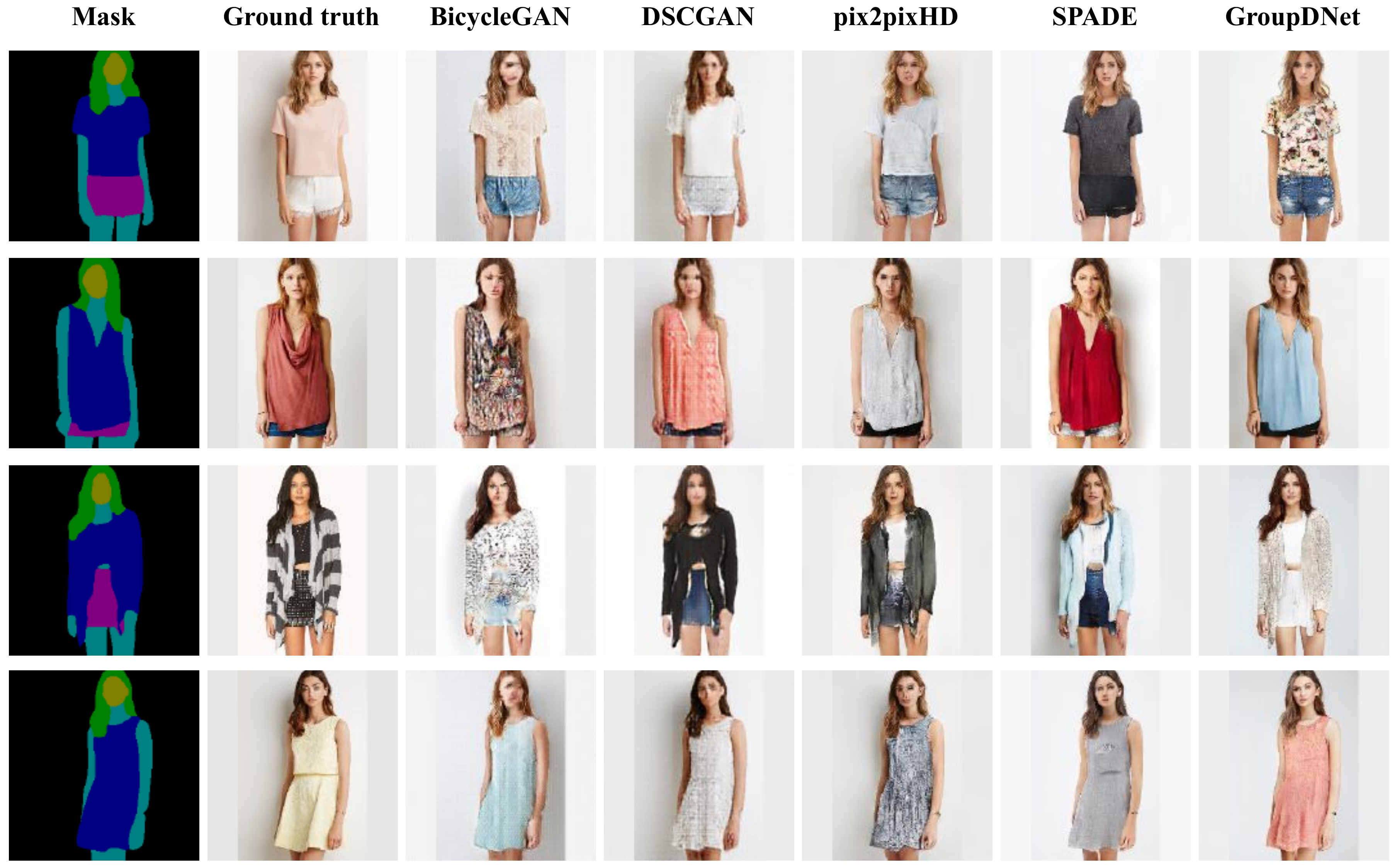}
	\caption{Qualitative comparison of our model with several label-to-image methods on the DeepFashion dataset.}
	\label{fig:deepfashion_comparison}
\end{figure*}
\begin{figure*}[tb]
	\centering
	\includegraphics[width=\textwidth]{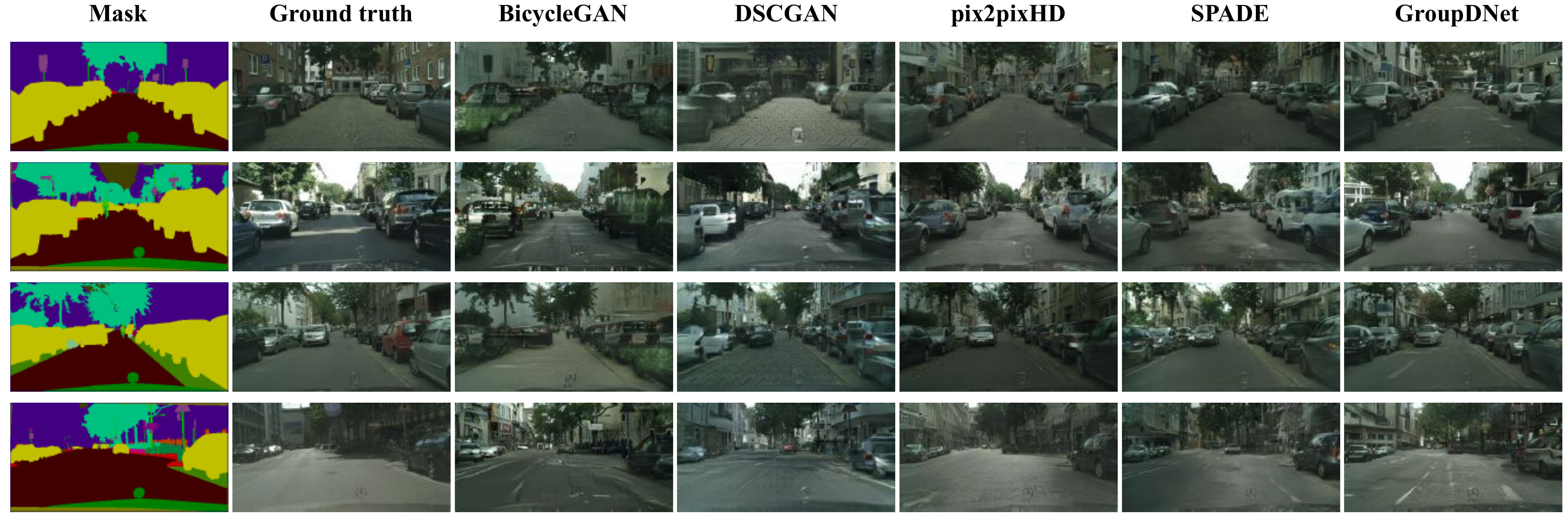}
	\caption{Qualitative comparison of our model with  several label-to-image methods on the Cityscapes dataset.}
	\label{fig:cityscapes_comparison}
\end{figure*}
\begin{figure*}[tb]
	\centering
	\includegraphics[width=\textwidth]{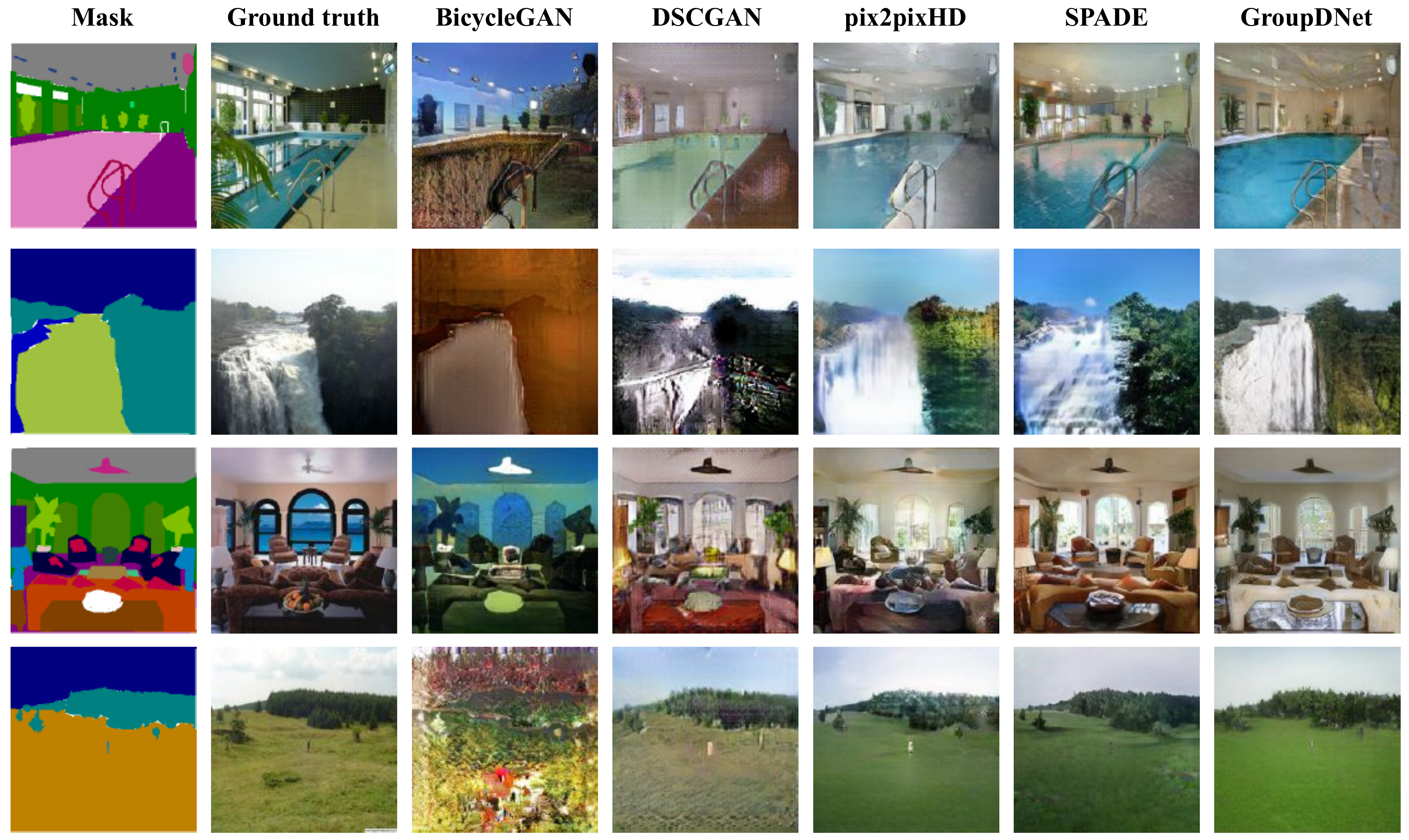}
	\caption{Qualitative comparison of our model with  several label-to-image methods on the ADE20K dataset.}
	\label{fig:ade20k_comparison}
\end{figure*}

In Fig~\ref{fig:deepfashion_comparison}, Fig~\ref{fig:cityscapes_comparison} and Fig~\ref{fig:ade20k_comparison}, we show additional comparison results from the proposed method on the DeepFashion, Cityscapes and ADE20K datasets with pix2pixHD~\cite{pix2pixHD} and SPADE~\cite{SPADE}. These results show that the image quality of GroupDNet is slightly better than the other two methods, especially in terms of keeping the object structures ordered and regular in the Cityscapes dataset (See the buildings and cars in these pictures).

In the accompanying video attached to our code base\footnote{\url{https://github.com/Seanseattle/SMIS}}, we demonstrate more results of our model on all datasets. Besides, we give a more straightforward demonstration of our exemplary applications. The video shows more results and detailed instructions of our exemplary applications in the main submission. From these videos, we exhibit the SMIS performance of GroupDNet on all three datasets and the potential applications of models designed for SMIS task. However, our model trained on the Cityscapes dataset seems to lose semantic controllability. For example, when altering the latent code for the buildings, other parts follow to change with the buildings. We are not willing to regard this phenomenon as a significant flaw.
In some cases, we do hope the whole image can change alongside the change of a class-specific latent code because it strengthens the fidelity of the generated images. For example, the discordance between the overall illumination and illuminations on some objects could make the image unrealistic and unnatural. Another problem is the diversity of the generated results on Cityscapes seems quite limited. It is because this dataset is restricted initially to the scenes of German cities.
Moreover, the images inside the dataset were shot during short intervals; hence, they exhibit no diversity of illumination from daylight to darkness. Seeing these results, we firmly believe semantically multi-modal image synthesis has more applications and intrinsic scientific values that deserve to explore given suitable datasets. In the future, we'll investigate more into GroupDNet and try to improve its performance in SMIS.

\section{Additional ablation study}

\begin{table}[tb]
	\centering
	\resizebox{0.48\textwidth}{!}{
		\begin{tabular}{|l|c|c|c|c|c|c|c|}
			\hline
			Models & FID$\downarrow$  & mCSD$\uparrow$ & mOCD$\downarrow$ & LPIPS$\uparrow$ & SHE$\uparrow$ & FPS $\uparrow$ & \# Para$\downarrow$  \\ \hline
			\textbf{GroupDNet} & \textbf{9.50} & 0.0264 & \textbf{0.0033} & 0.228 & \textbf{81.2} & 12.2 & 109.1 \\ \hline
			w/o map & 11.01 & 0.0253 & 0.0036 & 0.217 & 79.5 & 11.5 & 109.3 \\
			w/o split & 10.76 & 0.0054 & 0.0189 & 0.216 & 31.7 & 12.1 & 109.1 \\
			$\rightarrow$GroupNorm & 10.33 & 0.0256 & 0.0040 & 0.225 & 77.0 & 12.2 & 109.1 \\
			w/o SyncBN &9.76 & 0.0251 & 0.0037 & 0.216 & 79.3 & 12.3 & 109.1 \\
			w/o SpecNorm & 10.42 & \textbf{0.0290} & 0.0153 & \textbf{0.231} & 46.3 & \textbf{13.5} & 109.0 \\
			\hline
		\end{tabular}
	}
	\caption{Quantitative results of the ablation experiments on the DeepFashion dataset.}
	\label{tab:ablation}
\end{table}

To support for the SMIS task and improve the quality of generated images, we made several minor modifications to GroupDNet, including:
1) splitting the original input image to different images of different semantic classes, as mentioned in our main text;
2) enforcing the encoder to produce a mean map and variance map rather than a mean vector and variance vector that are used in SPADE. 
To validate the effects of these strategies, we conduct ablation experiments by not using them in the networks, thus we have the results of the model without splitting the original images (\textbf{w/o split}) and the model without producing a mean and variance map (\textbf{w/o map}). Note for the latter one, we use add global average pooling at the last of the encoder to compress maps into vectors, instead of using the routine fully connected layers to produce the vectors, which could cripple the individuality of each class. Results shown in Tab.~\ref{tab:ablation} indicates: splitting the input image or producing a mean and variance map are necessary strategies for the SMIS task, otherwise the model will suffer from a degraded performance of FID, mCSD and mOCD. 

Besides, considering the vast use of group convolution in GroupDNet, it is very interesting to know what the effect would be if we apply group normalization as the main normalization layer in our model because group normalization also operates separately on different groups of feature channels. Therefore, we conduct another experiment by changing all normalization layers in our original model to group normalization layers and set their group numbers equal to their previous convolutional layers (\textbf{$\rightarrow$GN}). We also conduct experiments to investigate the impacts of several normalization layers we use in our model by discarding the use of them. Basically, we then have the model without using synchronized batch normalization (\textbf{w/o SyncBN}) and without using spectral normalization~\cite{SpectralNorm} (\textbf{w/o SpecNorm}). Results are reported in Tab.~\ref{tab:ablation}. Changing the normalization layers to group normalization or removing the synchronized batch normalization have slightly degraded the performance on most metrics. However, removing the spectral normalization layers in GroupDNet will largely increase mOCD, indicating spectral normalization helps for the SMIS task. The LPIPS and mCSD metrics of the model without SpecNorm are even higher than GroupDNet, which hints that spectral normalization may have a negative effect on the diversity of the model's generated images.

\section{Discussions and future work}
As mentioned earlier, a limitation of GroupDNet is the restricted power to capture the class-wise modality of images in the Cityscapes dataset. Though the dataset itself demonstrates very few variation of object appearance, we believe there remains lots of methodological and architectural modifications that could enable GroupDNet or other models to handle such difficult cases.

Besides, we also feel the necessity to design a clean and effective strategy to decide how to set the group numbers for each convolution or possibly normalization layers. Though we haven't give clear experimental evidence, we feel that different configuration of group numbers might have some impacts on the performance. This conclusion is natural because different group number configurations determine different network structures and more often different network structures have influence on the network performance. 

Moreover, it is also interesting to discover whether changing the input order of different classes could make any difference to the performance, considering now we feed the split input images to the encoder, following the order set by the dataset provider or randomly set by us. It is quite natural to reason that putting similar classes together could make the corresponding areas change harmoniously and concurrently, thus producing images with much more fidelity.

\begin{table*}[tb]
	\centering
	\resizebox{\textwidth}{!}{
		\begin{tabular}{c|c|c|c}
			\hline
			Models & DeepFashion   & Cityscapes  & ADE20K   \\ \hline
			Encoder & $C=64,G=8$  & $C=8,G=35$  & $C=3,G=151$ \\
			Decoder & $C=160,G=\{8,8,4,4,2,2,1\}$  & $C=280,G=\{35,35,20,14,10,4,1\}$  & $C=\{151,64\},G=\{151,16,16,8,4,2,1,1\}$ \\ \hline
		\end{tabular}
	}
	\caption{Pre-defined hyperparameters of different datasets for our encoder and decoder. Note in Fig.~\ref{fig:encoder_arc}, Fig.~\ref{fig:decoder_fashioncity} and Fig.~\ref{fig:decoder_ade20k}, ``$C\{i\}$'' represent the $i$-th number inside the brace of $C$ and  ``$G\{i\}$'' likewise in the brace of $G$. }
	\label{tab:defined_numbers}
\end{table*}

\begin{figure*}[tb]
	\centering
	\includegraphics[width=\textwidth]{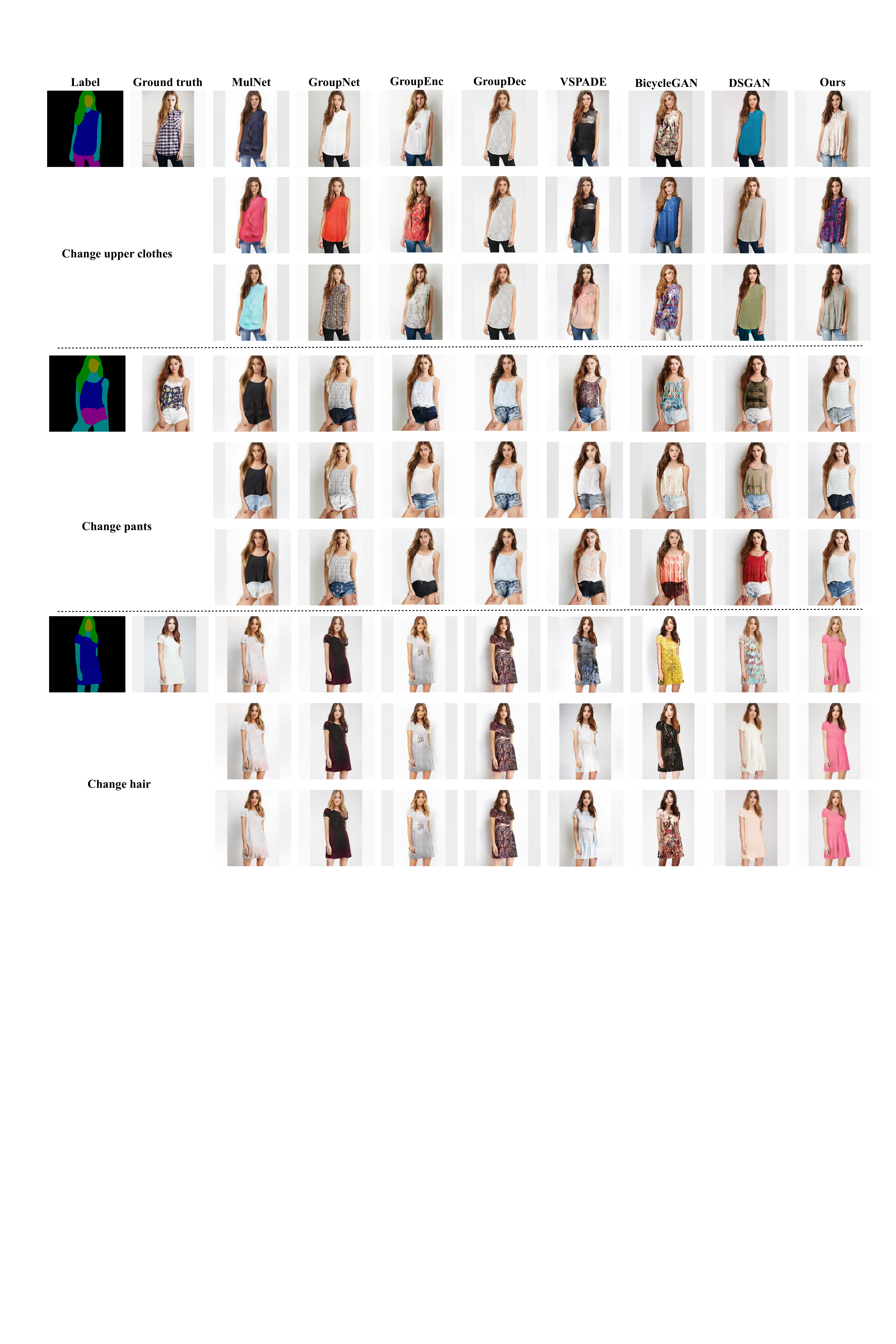}
	\caption{Qualitative comparison between GroupDNet and other baseline models. The first three rows represent the results of different models by changing their upper-clothes latent code. The middle three rows represent the results of different models by changing their pants latent code while the last three rows represent their results of changing the hair latent code. Note, for those models which have no class-specific latent code such as VSPADE, we alter their overall latent codes to generate different images.}
	\label{fig:ablation}
\end{figure*}

\end{document}